\pdfoutput=1

\documentclass[11pt]{article}

\usepackage{acl}

\usepackage{times}
\usepackage{latexsym}

\usepackage[T1]{fontenc}

\usepackage[utf8]{inputenc}

\usepackage{microtype}

\usepackage{textcomp}
\usepackage{makecell}
\usepackage{inconsolata}
\usepackage{url}
\usepackage{booktabs}
\usepackage{subcaption}
\usepackage{graphicx}
\usepackage{tabularx}
\usepackage{multirow}
\usepackage{multicol}
\usepackage{soul}
\usepackage{amsmath}
\usepackage{amssymb}
\usepackage{xstring}
\usepackage{xcolor}
\usepackage[noabbrev,capitalize,nameinlink]{cleveref}
\usepackage{color}

\newcommand{\escoxlmr}{\texttt{ESCOXLM-R}}
\newcommand{\promptbox}[1]{\begin{tcolorbox}[colback=yellow!25, colframe=black, rounded corners]
#1
\end{tcolorbox}}

\usepackage{awesomebox} 
\usepackage{bbding}
\usepackage[most]{tcolorbox}
\usepackage{fontawesome5}
\usepackage[tikz]{bclogo}
\usepackage[framemethod=tikz]{mdframed}
\usepackage[most]{tcolorbox}

\definecolor{bgblue}{RGB}{245,243,253}
\definecolor{ttblue}{RGB}{91,194,224}

\newif\ifcomments
\commentstrue
\ifcomments
\usepackage[normalem]{ulem}
\definecolor{ABpurple}{rgb}{0.8,0.0,0.8}
\newcommand{\ab}[1]{\textcolor{ABpurple}{\textsf{\scriptsize[\textbf{AB\@:} #1]}}} 
\newcommand\abi[1]{\textcolor{ABpurple}{#1}} 
\newcommand\abm[1]{\marginpar{\raggedright\tiny\textcolor{ABpurple}{\textsf{{\bfseries AB\@:} #1}}}} 
\newcommand\abs{\bgroup\markoverwith{\textcolor{ABpurple}{\rule[.4ex]{2pt}{0.8pt}}}\ULon} 
\else
\newcommand\ab[1]{}
\newcommand\abi[1]{\ignorespaces}
\newcommand\abm[1]{}
\newcommand\abs[1]{#1}
\fi

\newtcolorbox{prompt}{
  colback=openaigreen!15, 
  colframe=gray!60, 
  boxrule=1pt, 
  arc=0pt, 
  boxsep=0pt, 
  left=6pt, 
  right=6pt, 
  top=6pt, 
  bottom=6pt, 
  enhanced, 
  fontupper=\small,
  grow to left by=-1mm,
  grow to right by=-1mm,
  overlay={
    \node[anchor=north east] at (frame.north east) {\transparent{0.2}\includegraphics[width=8mm]{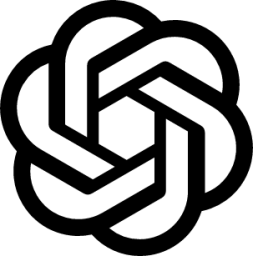}};
  }
}

\title{Rethinking Skill Extraction in the Job Market Domain using Large Language Models}

 \author{Khanh Cao Nguyen\textsuperscript{\faMountain} \hspace{2em}
         Mike Zhang\textsuperscript{\faCompass} \hspace{2em}
         \textbf{Syrielle Montariol}\textsuperscript{\faMountain} \hspace{2em}
         \textbf{Antoine Bosselut}\textsuperscript{\faMountain}\\
  \textsuperscript{\faMountain}EPFL, Switzerland\\
  \textsuperscript{\faCompass}IT University of Copenhagen, Denmark\\
{\tt syrielle.montariol@epfl.ch } \hspace{2em} {\tt mikejj.zhang@gmail.com}}
\begin{document}
\maketitle
\begin{abstract}
Skill Extraction involves identifying skills and qualifications mentioned in documents such as job postings and resumes. The task is commonly tackled by training supervised models using a sequence labeling approach with BIO tags. However, the reliance on manually annotated data limits the generalizability of such approaches. 
Moreover, the common BIO setting limits the ability of the models to capture complex skill patterns and handle ambiguous mentions. In this paper, we explore the use of in-context learning to overcome these challenges, on a benchmark of 6 uniformized skill extraction datasets. Our approach leverages the few-shot learning capabilities of large language models (LLMs) to identify and extract skills from sentences. We show that LLMs, despite not being on par with traditional supervised models in terms of performance, can better handle syntactically complex skill mentions in skill extraction tasks.\footnote{Code is available at \url{https://github.com/epfl-nlp/SCESC-LLM-skill-extraction} and data at \url{https://huggingface.co/datasets/jjzha}} 
\end{abstract}

\section{Introduction}

Skill Extraction (SE) is a challenging task in the job market domain that involves identifying and extracting specific skills mentioned in job postings, resumes, and other job-related documents.
SE plays a crucial role in various job market applications, such as matching job seekers with relevant job opportunities or analyzing trends in the job market. Prior approaches to SE rely on rule-based methods or keyword-matching techniques~\cite{khaouja2021unsupervised, ternikov2022soft}. More recent methods, which are considered state-of-the-art, propose to fine-tune language models to solve the task~\cite{zhang-etal-2022-skillspan, zhang-etal-2023-escoxlm}. However, they heavily rely on manually annotated data, which is prohibitively expensive to collect, especially in this application that requires human resource domain experts.

In this paper, we investigate the use of large language models (LLMs) for SE in the job market domain. LLMs have been trained on massive amounts of text data and have shown great potential in capturing the underlying patterns and semantics of language. The SE task can be linked with the more generic entity recognition task in natural language processing (NLP). Named Entity Recognition (NER) is a widely studied task in NLP that involves identifying and classifying named entities in text. The typical approach to the NER problem is to formulate it as a sequence labeling task that assigns each token to a predefined entity-related label. Recent methods have attempted to tackle this task using LLMs, with limited success (e.g.,~\citealp{wang2023gpt, ma2023large}).

First, we propose a review of the datasets for SE. We uniformize them and release them as a benchmark for future research in the field. It includes six publicly available datasets covering 4 languages (English, French, German and Danish) and various domains and skills categories: \textsc{Sayfullina} \cite{sayfullina2018learning}, \textsc{SkillSpan}~\cite{zhang-etal-2022-skillspan}, \textsc{Green}~\citep{green-etal-2022-development}, \textsc{Gnehm}~\cite{gnehm-etal-2022-fine}, \textsc{Kompetencer}~\cite{zhang-etal-2022-kompetencer} and \textsc{Fijo}~\cite{beauchemin-2022-fijo}.

We test the ability of LLMs to solve the tasks on these six datasets using various prompting techniques. We compare two types of task formulations, highlighting their impact on various aspects of the model's behavior. Through diagnosing the performance of LLMs, we identify a taxonomy of errors, quantifying their frequency and supporting it with descriptive statistics on the datasets and the model's predictions, comprising multiple skill mentions that are conjoined together. For example, a job posting might require ``ability to develop reporting software and statistical softwares". In this case, the phrase ``develop reporting software and statistical softwares" represents a conjoined skill, where multiple skills are combined together. However, the current BIO annotation scheme does not capture such complex skill mentions effectively, leading to difficulties in accurately extracting each skill. 
Following this, in our error analysis, we highlight LLM behavior that is seen as detrimental using common NER evaluation and annotation schemes, but may be more adapted to real-world settings.

\section{Related Work}

\subsection{Skill Extraction}

Recently, there has been an increase of interest in the task of SE.
The general dynamic nature of labor markets has led to an increase in tasks related to job descriptions (JD), including SE~\cite{kivimaki2013graph,zhao2015skill,sayfullina2018learning,bhola-etal-2020-retrieving,gugnani2020implicit,fareri2021skillner,konstantinidis2022knowledge, ao2023skill, zhang-etal-2023-escoxlm}. Some works define a more granular label space such as~\citet{zhang-etal-2022-skillspan}. Here, they distinguish between skill and knowledge components. In~\citet{sayfullina2018learning}, they only extract soft skills. Then, in~\citet{green-etal-2022-development}, they tag for both hard and soft skills. For other languages, in~\citet{gnehm-bhlmann-clematide:2022:LREC}, they classify for only ICT-based skills in German. Last, for~\citet{beauchemin-2022-fijo}, they tag for only soft skills in French.

All these works employ methods such as sequence labeling \cite{sayfullina2018learning, smith2019syntax, chernova2020occupational, zhang-etal-2022-skillspan, zhang2022skill}, multi-label classification on the document-level~\cite{bhola-etal-2020-retrieving}, and graph-based methods \cite{shi2020salience, goyal-etal-2023-jobxmlc}. 
Recent methodologies include domain-specific models where LMs are continuously pre-trained on unlabeled JD \cite{zhang-etal-2022-skillspan, gnehm-bhlmann-clematide:2022:LREC}.
However, to the best of our knowledge, no work has applied LLMs to the task of SE in the job market domain. We use several datasets from previous work in~\cref{sec:datasets}.

\subsection{Entity Recognition using LLMs}

With the recent advances in large-scale pre-training, LLMs were able to capture rich contextual information and achieve impressive performance in various downstream tasks \cite{DBLP:journals/corr/abs-2303-08774, DBLP:journals/corr/abs-2307-09288}. However, their performance in NER is still significantly below supervised baselines \cite{ma2023large, wang2023gpt}. This is partly because, as a sequence labeling benchmark, NER requires a structured output from models. Meanwhile, the autoregressive nature of LLMs does not guarantee a uniform output representation. \citet{jimenez-gutierrez-etal-2022-thinking} conducted a comprehensive study on entities and relation extraction using LLMs in the biomedical domain and concluded that contemporary techniques could not enable GPT-3 with in-context learning to surpass BERT-sized fine-tuned LMs. \citet{ma2023large} and  \citet{wang2023gpt} reach similar conclusions in the general domain. 
\citet{DBLP:journals/corr/abs-2304-10428} proposed a method to surround entities with special tokens, bridging the gap between sequence labeling and generative tasks and achieving comparable results compared to state-of-the-art pre-trained models. In more recent works, PromptNER \cite{DBLP:journals/corr/abs-2305-15444} provides the entity definition to the model, asking it to output a list of potential entities along with the reasoning on the compatibility of each entity with the provided definition.  Meanwhile, UniversalNER \cite{zhou2023universalner} instruction-tunes smaller scale open-source models for entity extraction tasks.

\section{Datasets for Skill Extraction}\label{sec:datasets}
We gather and uniformize six datasets with the B-I-O annotation scheme, where each word in a sentence is associated with one tag. The ``B'' tag indicates that the associated word marks the \textit{Beginning} of a span; ``I'' indicates a word \textit{Inside} a span, and ``O'' marks words \textit{Outside} a span. Each dataset is extracted from job ads from various domains and languages, and some are augmented with fine-grained annotations.
Despite the disparity of the train-dev-test split proportion, we kept the original splits from the authors of each dataset to ensure comparability with previously published results.

For each dataset, we describe how they were created, and their content. Table \ref{tab:datasets_overview} provides information on their size, language, domain and labels. Table \ref{tab:datasets_statistics} presents descriptive statistics of the six datasets, such as average sentence length and number of skills. 

\begin{table*}[!ht]
    \centering
    \resizebox{\linewidth}{!}{%
    \begin{tabular}{lrrrrrr}
    \toprule
        Name        & Train  & Dev   & Test  & Language      & Domain  & Labels \\
        \midrule
        \textsc{Green}       & 8,669  & 964    & 335   & English       & Multiple      & \texttt{Qualific., Domain, Occupation, Exp., Skills}\\
        \textsc{SkillSpan}   & 4,800  & 3,174  & 3,569 & English       & Multiple      & \texttt{Skills, Knowledge}\\
        \textsc{Sayfullina}  & 3,705  & 1,855  & 1,851 & English       & Multiple      & \texttt{Soft Skills}\\ 
        \textsc{Gnehm}       & 19,889 & 2,332  & 2,557 & German        & ICT           & \texttt{ICT}  \\
        \textsc{Fijo}        & 399    & 49     & 49    & French        & Insurance     & \texttt{Thoughts, Results, Relational, Personal}\\
        \textsc{Kompetencer} & 778    & 346    & 262   & Danish        & Multiple      & \texttt{Skills, Knowledge}\\
        \bottomrule
    \end{tabular}%
    }
    \caption{\textbf{Datasets Overview:} number of sentences in each split, language, the domain of the job ads, and which type of labels are originally present in the dataset.}
    \label{tab:datasets_overview}
\end{table*}

\begin{table*}[!ht]
    \centering
    \resizebox{\linewidth}{!}{
    \begin{tabular}{lrrrrr}
    \toprule
        Name        & Avg. Sentence Length  & Avg. \# of Skills   & \% Sentence w/o Skills & Avg. Skills Span Length      & Tot. Unique Skills  \\
        \midrule
        \textsc{Green}       & 22.94  & 2.0    & 30.45   & 2.68       & 610  \\
        \textsc{SkillSpan}   & 11.99  & 0.3  & 83.64 & 3.56  & 986   \\
        \textsc{Sayfullina}  & 14.35  & 1.0  & 0.1 & 1.75 & 581 \\ 
        \textsc{Gnehm}       & 10.77 & 0.3  & 82.82 & 1.32  & 675    \\
        \textsc{Fijo}        & 31.70 & 2.4  & 16.0  & 9.7  & 123   \\
        \textsc{Kompetencer} & 13.16 & 0.4 & 83.2 & 3.79  & 103  \\
        \bottomrule
    \end{tabular}}
    \caption{\textbf{Test Split Datasets Statistics:} Average sentence length (number of words), average number of labeled skills per sentence, percentage of sentences in the dataset without any skill, average number of words in a skill span, and total number of unique skills in the test set.}
    \label{tab:datasets_statistics}
\end{table*}

\paragraph{\textsc{Fijo}~\citep{beauchemin-2022-fijo}}\footnote{\url{https://huggingface.co/datasets/jjzha/fijo}} 
A French job ad dataset annotating skill types using a sequence labeling approach. The skill groups are based on the AQESSS public skills repositories and proprietary skill sets provided by their collaborators. These skill types are divided into four categories: ``Thoughts'', ``Results'', ``Relational'', and ``Personal''. 

\paragraph{\textsc{Gnehm}~\citep{gnehm-etal-2022-fine}}\footnote{\url{https://huggingface.co/datasets/jjzha/gnehm}} 
This is a Swiss-German job ad dataset focusing on Information and Communications Technology (ICT)-related sequence labeling. It includes ICT tasks, technology stack, responsibilities, and so forth. This dataset is a combination of two other Swiss datasets, the Swiss Job Market Monitor and an online job ad dataset~\cite{gnehm2020text, buchmann2022swiss}. 

\paragraph{\textsc{Green}~\citep{green-etal-2022-development}}\footnote{\url{https://huggingface.co/datasets/jjzha/green}} 
This English dataset is a token-level sequence labeling task containing five types of tags: Skills, Qualifications, Domain, Experience, and Occupation labels. The JDs present in the dataset are from the United Kingdom. The industries represented in the data are from various fields, such as IT, finance, healthcare, and sales.

\paragraph{\textsc{Sayfullina}~\citep{sayfullina2018learning}}\footnote{\url{https://huggingface.co/datasets/jjzha/sayfullina}} 
This dataset, in English, focuses on soft skill prediction. Soft skills are personal qualities such as ``team working'', ``being dynamic'', and ``independent''.

\paragraph{\textsc{SkillSpan}~\citep{zhang-etal-2022-skillspan}}\footnote{\url{https://huggingface.co/datasets/jjzha/skillspan}} 
This job posting dataset includes annotations for skills and knowledge, derived from the ESCO taxonomy. In a nutshell, \emph{knowledge} are things that you can learn, and \emph{skills} show how you apply this knowledge. The skills can be relatively long. The data is in English and contains JDs from Stackoverflow and a source that contains various other positions.

\paragraph{\textsc{Kompetencer}~\citep{zhang-etal-2022-kompetencer}}\footnote{\url{https://huggingface.co/datasets/jjzha/kompetencer}} 
This dataset includes JDs in Danish. The annotation scheme is the same as \textsc{SkillSpan}.

\section{Method}
In this work, we formulate the task of skills extraction as a LLMs generation task, leveraging LLMs' emergent in-context learning ability. The LLM is provided instructions, a set of demonstrations, and a sample to annotate. The instructions define the expected output format precisely. This format is crucial for NER tasks, as the rigid structure of the annotations and metrics poses a challenge in evaluating the generative output. In the following section, we describe our approaches and strategies to design prompt formats.

\subsection{Prompting Strategies}
\begin{figure}[t]
  \centering
  \includegraphics[width=0.5\textwidth]{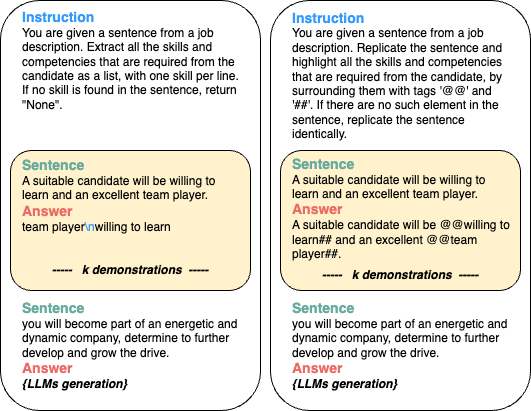}
  \caption{\textbf{Prompting Approaches.} \textsc{Extraction-style} \textbf{(left)}: The model extracts skills and presents them as a list, joined by a separator token; \textsc{Ner-style} \textbf{(right)}: The LLM rewrites the original sentence, with all skill mentions wrapped by special tokens ``\textit{@@}'' and ``\textit{\#\#}''.}
  \label{fig:prompt}
\end{figure}

We investigate two prompting strategies: \textsc{Extraction-style} and \textsc{Ner-style}, illustrated in Figure \ref{fig:prompt}. 

\paragraph{\textsc{Extraction-style}} The spans extracted from the sentences are directly generated as a list, as shown in Figure \ref{fig:prompt}. The output format is generated from the  BIO-tag annotations by extracting the skills and concatenating them with a separator (in this case, a ``\textit{\textbackslash n}'' token).

\paragraph{\textsc{Ner-style}} 
Following \cite{wang2023gpt}, we formulate the output format by rewriting the original sentence and adding special tokens around each entity. This leads to a more constrained output format, leaving less space for hallucination. In 
 practice, we wrap every skill from the original sentence with special token \textit{"@@"} and \textit{"\#\#"}, as shown in Figure \ref{fig:prompt}. 

 \paragraph{Dataset-Specific Prompt}
 In the absence of supervision from a large train set, the model has no way of knowing what exactly are the entities it is asked to extract. We create dataset-specific prompts to guide it, highlighting key information about the input sentence (domain, language) and the targeted entities (type of skills). We follow a specific template:

\promptbox{\texttt{You are given a sentence from a job description in <LANGUAGE>, in the <DOMAIN> domain. Extract all the <SKILL TYPE> that are required from the candidate, <PROMPT-SPECIFIC OUTPUT>.}}

\noindent
For example, for the \textsc{Gnehm} dataset with \textsc{Extraction-style} prompt, we have: 

\promptbox{\texttt{You are given a sentence from a job advertisement in German. Extract all the IT/Technology skills and competencies that are required from the candidate as a list.}}

\subsection{Demonstration Selections}
LLMs are sensitive to the different combinations of in-context examples \cite{liu-etal-2022-makes, wang2023large}. We experiment with two strategies to select demonstrations.
\paragraph{Semi-random Demonstrations} We randomly select $k$ examples from the train set. Due to the noisiness of the data, we manually inspect the examples, excluding low-quality ones. This process is time-consuming for high values of $k$; however, in our setting, having a high number of demonstrations ($k>5$) doesn't improve the extraction performance (see Figure \ref{fig:app-fewshots} in Appendix).

\paragraph{$k$NN-retrieval demonstrations} To leverage demonstrations that are closely related to each sample, we use a $k$NN-retrieval approach. We embed each sentence using in-domain, monolingual masked language models (MLM), and retrieve the top $k$ closest sentences in the train set using cosine similarity. 
\paragraph{Mix of Positives and Negatives} We define a \textit{\textbf{negative}} demonstration as a sample from the training subset that does not contain any skill, and a \textit{\textbf{positive}} one otherwise. 

We create a mix of positive and negative demonstrations with a 1:1 ratio. When performing $k$NN-retrieval, we retrieve the nearest neighbors separately in the pools of positive and negative examples. Throughout our experiments, we denote \textit{k}-shot a prompt with $k$ positives and $k$ negatives demonstrations.

\subsection{Post-processing} Depending on the prompting strategies, we implement post-processing pipelines to extract the entities from the LLM's outputs. 
Indeed, since LLMs are trained for text generation instead of sequence labeling, they struggle with replicating spans of the input sequence as required for NER.

Even with in-context examples, LLMs fail to generate the required output correctly. In particular, on top of failing to respect the format (e.g., adding the right tags around entities for the \textsc{NER-style prompting}), it often fails to correctly replicate the spans of the input sentence. Indeed, since the model generates the most probable sequence, it usually attempts to correct the errors found in the input sentence, such as spacing around punctuation or typos. Thus, in the process, it modifies the initial spans and prevents them from being accurately matched with the original input sentence, hindering the evaluation. 

To address this issue, we propose a rule-based post-processing step that handles mismatched punctuation and minor edits made to the span by the model. We manually look at a sample of mismatches between LLM generations and original sentences and identify a set of common mismatches (see examples in Appendix \ref{ap:mismatches} for categories and examples). We automatically detect if the model generation falls in one of these cases, and correct it using heuristics (e.g. added/removed spaces, punctuation insertion or deletion). Otherwise, we implement a feedback loop to prompt it to correct its answer. We feedback on the original prompt as well as the model's answer, describe the mistake it makes, and request another generation with the instructions below. \\

\noindent
\textsc{Extraction-style}: 

\promptbox{\texttt{You have correctly extracted these skills: <CORRECTLY EXTRACTED SKILL>.
The following skills you extracted are either absent or not written the same way as in the original sentence: <INCORRECTLY EXTRACTED SKILL>. Modify these skills to make sure to exactly replicate these skills from the input sentence with their original spellings and grammars, discard any of them if needed. Remember to keep the skills that you correctly extracted.  Provide them with one skill per line.}}

\noindent
\textsc{NER-style:} 

\promptbox{\texttt{You didn't correctly replicate the given sentence. Make sure the sentence stays the same, even if there are no skills to highlight, including punctuation, spacing, and grammar mistakes. Don't add any extra words or punctuation to the sentence except for the \#\# and @@ tags. Don't add nor remove any space. Remember to keep the valid highlighted skills with tags '@@' and '\#\#': <CORRECTLY EXTRACTED SKILL>}}

If one feedback loop is not enough, we repeat the process of providing feedback and requesting another generation up to 3 times. If the desired format is not achieved after 3 retries, we consider that the model failed to extract any entity. We perform an extensive analysis of failure cases in Section \ref{sec:error}.\\
\section{Experiments}

\subsection{Experimental Framework}
\paragraph{Models}
We use GPT-3.5-turbo\footnote{\texttt{gpt-3.5-turbo-instruct}} for all of our experiments, which has an input context of 4096 tokens. We set the temperature to 0 
to enforce deterministic generation from the model. We also experiment with GPT-4 to set an upper-bound expectation for the performance. Due to budget constraints, we evaluate GPT-4 on a subset of samples.

To retrieve demonstrations, we use monolingual pre-trained models adapted to each dataset language. If possible, we use models fine-tuned on domain-specific datasets: JobBERT\footnote{\url{https://huggingface.co/jjzha/jobbert-base-cased}} for English, DaJobBERT\footnote{\url{https://huggingface.co/jjzha/dajobbert-base-uncased}} for Danish, jobBERT-de\footnote{\url{https://huggingface.co/agne/jobBERT-de}} for German, and CamemBERT\footnote{\url{https://huggingface.co/camembert-base}} for French. The latter is the only one which is not specifically fine-tuned on jobs postings.

\paragraph{Baselines}
As baselines, we use the supervised results from~\citet{zhang-etal-2023-escoxlm} which are currently state-of-the-art. The model is \escoxlmr{}, an \texttt{XLM-R}$_\text{large}$-based encoder model~\cite{conneau2020unsupervised}, further pre-trained using the ESCO taxonomy~\cite{le2014esco} by employing a combination two training objectives: Masked language modeling and a three-way classification of whether concepts in ESCO are in connection with each other
, to adapt the model to the job market domain. The supervised results are from fine-tuning the model on the training set of each dataset separately. 

\paragraph{Evaluation Metrics}
The metrics are Precision (\textbf{P}), Recall (\textbf{R}) and span-F1 (\textbf{F1}).
We compute \textsc{Strict} metrics using \texttt{seqeval}.\footnote{\url{https://github.com/chakki-works/seqeval}} We implement a \textsc{Relax} skill-level metric, in which we considered an extracted entity as correct even if it only \textit{\textbf{partially overlaps}} with the gold span from the annotation. The \textsc{Relax} metric aims to evaluate the ability of LLMs to localize the skills within the given sentence.

\subsection{Experimental Results}

\begin{table*}[t]
    \centering
    \footnotesize
    \resizebox{\linewidth}{!}{
    \begin{tabular}{l *{1}{c} *{1}{c} *{1}{c} *{1}{c} *{1}{c} *{1}{c} *{3}{c}}
        \toprule
        & \multicolumn{1}{c}{\textbf{\textsc{Fijo}}}  
        & \multicolumn{1}{c}{\textbf{\textsc{Gnehm}}} 
        & \multicolumn{1}{c}{\textbf{\makecell{\textsc{Kompe-}\\ \textsc{tencer}}}} 
        & \multicolumn{1}{c}{\textbf{\textsc{Green}}}  
        & \multicolumn{1}{c}{\textbf{\makecell{\textsc{Skill-}\\ \textsc{Span}}}} 
        & \multicolumn{1}{c}{\textbf{\makecell{\textsc{Sayful-}\\ \textsc{lina}}}} 
        & \multicolumn{3}{c}{\textbf{\textsc{Avg}}}\\
        \cline{8-10}
        \textsc{STRICT} & F1 & F1 & F1 & F1 & F1 & F1 & P & R & F1\\
        \midrule
        \textsc{Supervised} & 42.0 & 88.4 & 49.8 & 51.2 & 62.6 & 92.2 & -- & -- & 64.4\\
        \midrule
        \textsc{Extract-Style} &  &  &  &  &  &  & \\
        \hspace{0.5em}zero-shot+\textit{specific} & 0.0 & 21.0 & 15.9 & 4.4 & 6.5 & 11.6 & 19.8 & 7.15 & 9.9\\
        \hspace{0.5em}5-shot & 28.7 & 27.5 & 21.6 & 24.2 & \textbf{25.0} & 29.3 & 23.3 & 35.5 & 26.1\\
        \hspace{1em}\textit{+$k$NN} & 34.3 & 29.0 & \textbf{22.4} & 29.0 & 20.9 & 33.3 & 24.4 & 39.4 & 28.1\\
        \hspace{1em}\textit{+$k$NN+specific} & 35.7 & 40.5 & 20.9 & 28.4 & 20.3 & \textbf{39.0} & \textbf{26.6} & \textbf{41.1} & \textbf{30.8}\\
        \midrule
        \textsc{Ner-style} &  &  &  &  &  &  & \\
        \hspace{0.5em}zero-shot+\textit{specific} & 3.0 & 7.4 & 1.6 & 0.7 & 2.3 & 0.4 & 6.53 & 1.8 & 2.57\\
        \hspace{0.5em}5-shot & 33.3 & 33.1 & 20.4 & 28.7 & 17.8 & 27.0 & 23.8 & 35.5 & 26.7\\
        \hspace{1em}\textit{+$k$NN} & 36.7 & 32.3 & 15.3 & \textbf{32.0} & 15.5 & 32.3 & 23.3 & 39.3 & 27.4\\
        \hspace{1em}\textit{+$k$NN+specific} & \textbf{44.2} & \textbf{40.9} & 16.1 & 31.8 & 13.7 & 36.6 & 26.4 & 40.9 & 30.5\\
        \midrule
        \textsc{GPT-4*} &  &  &  &  &  &   & \\
        \hspace{0.5em}\textsc{Extract-style} & 38.0 & 58.7 & \textbf{25.3} & \textbf{30.6} & \textbf{27.8} & \textbf{40.5} & 33.3 & \textbf{44.1} & 36.8\\
        \hspace{0.5em}\textsc{Ner-style} & \textbf{48.0} & \textbf{67.8} & 24.6 & 21.9 & 25.7 & 38.4 & \textbf{35.8} & 42.7 & \textbf{37.7}\\
        \midrule
        \midrule
        \textsc{RELAX} & F1 & F1 & F1 & F1 & F1 & F1 & P & R & F1\\
        \midrule
        \textsc{Extract-Style} &  &  &  &  &  &   & \\
        \hspace{0.5em}5-shot & 70.0 & 35.0 & \textbf{46.3} & 70.1 & 45.7 & 48.9 & 48.5 & 67.9 & 52.7\\
        \hspace{1em}\textit{+$k$NN} & 80.3 & 35.6 & 45.9 & \textbf{74.0} & \textbf{46.2} & 50.6 & 49.1 & 73.7 & 55.4\\
        \hspace{1em}\textit{+$k$NN+specific} & 78.8 & 47.8 & 44.8 & 73.6 & 45.4 & \textbf{57.7} & \textbf{51.1} & \textbf{75.1} & \textbf{58.0}\\
        \midrule
        \textsc{Ner-style} &  &  &  &  &  &   & \\
        \hspace{0.5em}5-shot & 57.4 & 28.9 & 37.0 & 64.1 & 32.6 & 47.6 & 41.8 & 59.9 & 44.6\\
        \hspace{1em}\textit{+$k$NN} & \textbf{82.0} & 37.9 & 33.2 & 64.7 & 29.9 & 48.2 & 42.8 & 68.5 & 49.3\\
        \hspace{1em}\textit{+$k$NN+specific} & 80.3 & \textbf{48.2} & 44.8 & 66.0 & 27.6 & 51.4 & 46.5 & 69.3 & 53.1\\
        \bottomrule
    \end{tabular}}
    \caption{\textbf{Results of Experiments}, measured using Precision (\textbf{P}), Recall (\textbf{R}) and span-F1 (\textbf{F1}) metrics. \textit{*GPT-4 results are based on a subset of  $\leq350$ samples for each dataset.}}
    \label{tab:results}
\end{table*}

 Table \ref{tab:results} compare the two prompting styles with various demonstration retrieval settings: zero-shot, 5-shots (a mix of 5 randomly retrieved negative demonstrations\footnote{Except for \textsc{Sayfullina}, where there are only 4 negative training samples.} and 5 random positive demonstrations), and 5-shots+$k$NN (top 5 nearest neighbors retrieved from the set of negative examples, and top 5 positive examples). All examples are retrieved from the train set of the datasets. The choice of 10 demonstrations stems from an ablation study reported in Figure \ref{fig:app-fewshots} in Appendix. In the zero-shot setting, we always use the dataset-specific prompts to guide the model toward the desired type of entities to extract. The full table with precision and recall can be found in appendix (Table \ref{tab:results-full}).

There is a large drop in performance across most datasets compared to fine-tuning models, for both the \textsc{Extract} and \textsc{Ner-style} approaches. Aside from \textsc{Fijo}, in which GPT-3.5 with in-context learning achieves a comparable performance, in all other datasets the decrease margin is significant, with up to 50\% decreases in F1 metrics. 

The datasets with the largest performance drop are \textsc{Gnehm} and \textsc{Sayfullina}. We hypothesize that pre-trained models achieve better performances when the length of span entities is rather short (Table \ref{tab:datasets_statistics}). 
The \textsc{Relax} metric shows much higher performance (up to 20\% higher average F1 score). In particular the recall is considerably higher, showing that LLMs are able to localize the skills within a sentence, but fail to capture the exact sequence. Concisely, we have the following findings:

\noindent
\textbf{Few-shot demonstrations are critical to model performance,~} with an average improvement of 20.0\% for \textsc{Extract-style} and 28\% for \textsc{Ner-style} in F1 when providing 5-shot demonstrations. Indeed, given the strict structure required by NER, it is essential to show the model examples of the specific output format, especially for \textsc{Ner-style} format.

\noindent
\textbf{\textsc{Extract-style} outperforms \textsc{Ner-style} prompting on average}, especially with the \textsc{Relax} evaluation scheme, in contrast to the findings of \citet{
DBLP:journals/corr/abs-2304-10428}. 

\noindent
\textbf{$k$NN-retrieval outperforms random selection of demonstrations} across all benchmarks except \textsc{Skillspan}. On average, switching from random to $k$NN demonstrations slightly improves the precision but greatly improves the recall (4\% increases for both \textsc{Extract-style} and \textsc{Ner-style}). 
\noindent
\textbf{Dataset-specific prompts improve the performance} on average, particularly on datasets highly specialized toward a specific domain (insurance for \textsc{Fijo}, IT for \textsc{Gnehm}) or skill type (soft skills only for \textsc{Sayfullina}). 

\subsection{Error analysis}

\label{sec:error}
\begin{table}[t]
    \centering
    \resizebox{\linewidth}{!}{
    \begin{tabular}{lrr}
    \toprule
    Strategies & \textsc{Extract-style} & \textsc{Ner-style} \\
    \midrule
    zero-shot\textit{+specific} & 26.5162 & 12.8629 \\
    5-shot & 3.4623 & 2.2982 \\
    \hspace{0.5em}\textit{+$k$NN} & 1.8105 & 3.1458 \\
    \hspace{0.5em}\textit{+$k$NN+specific} & 1.6867 & 1.9816 \\
    \bottomrule
    \end{tabular}}
    \caption{Average percentage of samples for which the LLM failed to extract entities after 3 re-tries. Results are averaged across all datasets. The zero-shot setting gave the highest number of failures. }
    \label{tab:failures}
\end{table}

\begin{figure*}[t]
  \centering
  \begin{subfigure}[b]{0.3\textwidth}
    \includegraphics[width=\textwidth]{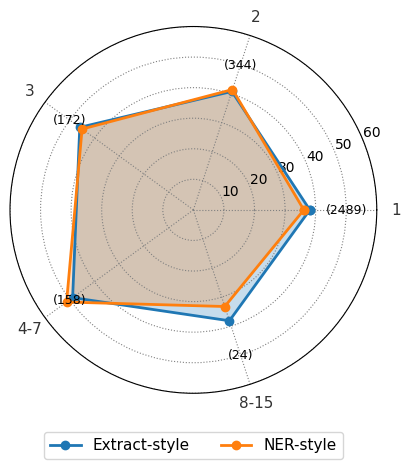}
    \caption{\textit{Number of skills}}
    \label{fig:sub1}
  \end{subfigure}
  \hfill
  \begin{subfigure}[b]{0.3\textwidth}
    \includegraphics[width=\textwidth]{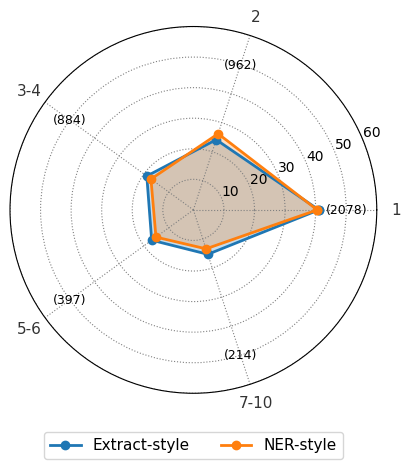}
    \caption{\textit{Skills span length}}
    \label{fig:span_len}
  \end{subfigure}
  \hfill
  \begin{subfigure}[b]{0.3\textwidth}
    \includegraphics[width=\textwidth]{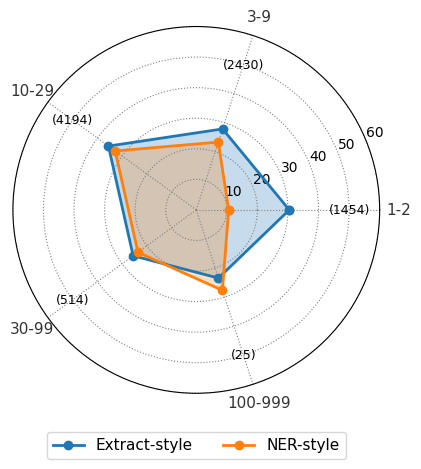}
    \caption{\textit{Sentence length}}
    \label{fig:sent_len}
  \end{subfigure}
  \caption{\textbf{Analysis Statistics.} The F1 scores vary across different criteria. \textbf{(left)} Extraction performances across different numbers of skills in the sentence (excluding negative samples). \textbf{(middle)} Extraction performances for each bucket of lengths of skills span. \textbf{(right)} Extraction performances for different ranges of the sentences' length. We used the number of tokens as the measure of length. For simplicity, this error analysis was performed on the set of all datasets.}
  \label{fig:error_analysis}
\end{figure*}

\paragraph{Failure cases related to instruction-following errors} Despite our post-processing pipeline, the LLM output often drifts from the desired format. Table \ref{tab:failures} shows the number of failure cases from different prompting strategies; a failure case happens when the LLM fails to output the desired format after 3 feedback loops. \textit{Zero-shot} inference exhibits the largest number of failures across almost all experiments by a large margin, as the instructions are not detailed enough to cover all potential output format divergences by the LLM. Providing demonstrations drastically reduces the number of failure cases. However, using $k$NN-retrieval does not necessarily reduce it further, even increasing the failure rate for \textsc{Ner-style}. 

\paragraph{Impact of various dataset features} 
Figure \ref{fig:error_analysis} shows how several sample features affect the extraction performances of the LLM, for both prompting strategies. \textbf{The number of skills per sentence} does not greatly affect extraction capabilities, for both \textsc{Extraction-style} and \textsc{Ner-style}. 
The \textbf{Skill span length} (from 1 to 10 words) affects the performance: entities with short span length, typically 1-2 words, are more likely to be correctly extracted. F1 performance degrades as the span gets longer.
Finally, the \textbf{sentence length} heavily affects \textsc{Ner-style} performance, which is considerably lower for short sentences (1-2 words) and gradually improves with longer ones. Meanwhile, \textsc{Extract-style} keeps a relatively stable performance across different sentence lengths. 

\paragraph{General Behavior}

Table \ref{tab:differences} showcases the flaws in the skills extraction task performed by GPT-3.5. Compared to the ground truth annotations, the LLM extracts shorter spans (on average over all datasets, 2.1 fewer words for \textsc{Ner} prompting) and more skills (on average over all datasets, 0.33 more skills per sample for both prompting strategies). To diagnose the reasons for this discrepancy, we analyze a sample of errors.

\subsection{Error taxonomy}

\begin{table}[!t]
    \centering
    \begin{tabular}{lrr}
    \toprule
    Prompt & Span Length & \# Skills  \\
    \midrule
    \textsc{Ner} & -2.10 & +0.33 \\
    \textsc{Extract} & -1.46 & +0.31 \\
    \bottomrule
\end{tabular}
    \caption{Difference between skills extracted by GPT-3.5 and gold annotations, for each prompting strategy.\textit{ Interpretation: GPT-3.5 extracts on average 0.3 more skills than the gold annotation.} We use $k$NN-5-shots with dataset-specific prompts. Full results per dataset can be found in Table \ref{tab:app-differences} in Appendix.}
    \label{tab:differences}
\end{table}

We manually analyze 60 examples where LLMs made the wrong predictions, both with \textsc{Extract-style} or \textsc{Ner-style}. The examples are randomly extracted with 10 from each dataset. Based on the analysis, we clustered the types of errors/misalignments between predictions and ground truths into different categories, listed below. Note that one sample can belong to several categories. We provide examples of each category in Appendix \ref{ap:qual_analysis} (Table \ref{tab:qual_analysis}):

\begin{enumerate}
    \item \textbf{Skill definition mis-alignment.~} Misalignments between the definition of skills used by human annotators in the ground truth data, and what is considered a skill by the LLM. The errors are often career-related terminologies that are not exactly skills or competencies, but are still extracted by the LLM. This category accounted for up to $\sim$36\% of the examined examples. We hypothesize that this category requires more demonstrations and more detailed instructions, potentially including definitions of skill types. The supervised model suffers much less from this flaw.
    \item \textbf{Wrong extraction.~} At times, the LLM also extracts spans that are completely unrelated to skills, competencies, or other career-related terminologies. 20\% of samples fall into this category. 
    \item \textbf{Conjoined skills.~} Here, we describe cases where a common span encompasses two skills. As an example, a job posting might require the "ability to develop reporting software and statistical software". The phrase "develop reporting software and statistical software" is a conjoined skill, where two skills (develop reporting software, and develop statistical software) are combined into a unique verb phrase. $\sim$14\% of errors are attributed to this category. We approximate the true number of conjoined skills in the annotations for each dataset, using syntactic parsing. According to our measure, 16 to 22\% of spans in 4 out of 6 datasets are conjoined skills. This observed behavior is in line with the metrics computed in Table \ref{tab:differences} on differences between the LLM predictions and the ground truth annotations. Interestingly, we note that the prevailing BIO annotation scheme can not distinguish such skill mentions by construction, merging them into a unique span and yielding errors as well.
    \item \textbf{Extended span.~} $\sim$12\% of the errors were instances where the LLM extracted longer spans than the ground truth, either by concatenating distinct skill mentions, or, more frequently, taking additional words around the gold skill mention. 
    \item \textbf{Incorrect annotations.~} 8\% of the errors were due to poor gold annotations. The annotated entity, while perhaps related to job markets, is irrelevant to skills, given the context. 
    \item \textbf{Other.~} Other unidentified minor cases, such as LLM generations that do not conform to the expected format ($\sim$10\%). A common cause is the grammatical correctness of the input sentence, to which LLMs are extremely sensitive. In cases where the original sentence has an error, the LLM is attempting to correct it while solving the task. Consequently, when performing the evaluation, the generation from the model cannot be exactly matched with the original sentence.
\end{enumerate}

\begin{table}[t]
    \centering
    \begin{tabular}{lr}
    \toprule
    Dataset & \% Conjoined skill \\
    \midrule
    \textsc{green} & 21.87 \\
    \textsc{skillspan} & 22.39 \\
    \textsc{fijo} & 17.79 \\
    \textsc{sayfullina} & 4.27 \\
    \textsc{kompetencer} & 16.67 \\
    \textsc{gnehm} & 0.02 \\
    \bottomrule
    \end{tabular}
    \caption{Proportion of entities that are conjoined skills in each dataset, obtained by performing syntactic parsing on the sentence.}
    \label{tab:conj_skills_percent}
\end{table}

\section{Conclusion}

In this paper, we benchmark and uniformize existing datasets for SE in job postings. We conduct in-depth experiments and error analysis to evaluate the ability of LLMs to solve the task, notably implementing two prompting strategies to adapt LLMs for the task and a dedicated feedback loop. In line with concurrent work \cite{han2023information}, LLMs achieve limited performance for skill extraction relative to supervised methods. Moreover, we highlight the limitations of the current SE task formulation and evaluation, focusing on the adaptation of the NER sequence labeling task, to the token generation task with which LLMs are pre-trained. 

In particular, we list the causes of the most frequent errors in SE with GPT-3.5. In the absence of training data, LLMs struggle to understand what skills are and often extract irrelevant information. Additionally, GPT-3.5 tends to split conjoined skills into two, leading to less accurate but more granular skill extractions. In a real-world setting, in particular when SE is used as a preliminary step for skill classification in a taxonomy (e.g. ESCO, \citealp{le2014esco}), this behavior would be highly beneficial.

\section{Limitations}

Several limitations to this study should be considered.

\noindent\textbf{Language.}
Despite our attempt to include as many datasets as possible, we are still limited in terms of language, as we include only four relatively high-resource languages: English, German, French, and Danish. This limits the generalizability of our findings to other languages, both in terms of the performance of LLMs and broader conclusions on the SE task.

\noindent\textbf{Closed-source Models.}
Our analysis is done exclusively using closed-source models, gpt-3.5-turbo and gpt-4. We have little information on the data, architecture, and training processes of these models, which heavily limits our ability to interpret and justify their performance. Similarly, these models are updated regularly, limiting the reproducibility of these results. 

\noindent\textbf{Biases.}
Pre-trained language models suffer from the bias present in their training data and reflect it in their predictions. While state-of-the-art supervised models for SE are also pre-trained models (e.g. ESCOXLM-R \cite{zhang-etal-2023-escoxlm}, fine-tuned from XLM-R), fine-tuning them on a high-quality, bias-controlled dataset for SE can mitigate the inherent bias present in the pre-training data \cite{wang2023overcoming}. Controlling the bias this way in LLMs such as GPT-3.5, when used in an in-context learning setting, is not an option and remains a challenging problem \cite{gallegos2023bias}. Biased models, when implemented in the job market domain, can have serious downstream consequences on the hiring process of candidates, particularly with respect to under-represented communities.

\section*{Acknowledgements}

We thank Jibril Frej (EPFL) for fruitful discussions and feedback on the first version of the paper. MZ is supported by the Independent Research Fund Denmark (DFF) grant 9131-00019B and in parts by ERC Consolidator Grant DIALECT 101043235. We also gratefully acknowledge the support of the Swiss National Science Foundation (No. 215390), Innosuisse (PFFS-21-29), the EPFL Science Seed Fund, the EPFL Center for Imaging, Sony Group Corporation, and the Allen Institute for AI.

\bibliography{anthology,custom}

\appendix

\section{Additional Analysis}
\label{app:results}

\paragraph{Main Results}
In addition to the F1 score, we also provide detailed results including the precision and recall for each prompting strategy, for both \textsc{Strict} and \textsc{Relax} metrics. Results are in Table \ref{tab:results-full}.

\begin{table*}[t]
    \centering
    \resizebox{\linewidth}{!}{
    \begin{tabular}{l *{3}{c} *{3}{c} *{3}{c} *{3}{c} *{3}{c} *{3}{c} c}
        \toprule
        & \multicolumn{3}{c}{\textbf{\textsc{Fijo}}}  & \multicolumn{3}{c}{\textbf{\textsc{Gnehm}}} & \multicolumn{3}{c}{\textbf{\textsc{Kompetencer}}} & \multicolumn{3}{c}{\textbf{\textsc{Green}}}  & \multicolumn{3}{c}{\textbf{\textsc{SkillSpan}}} & \multicolumn{3}{c}{\textbf{\textsc{Sayfullina}}} & \textbf{\textsc{Avg}}\\
        \textsc{STRICT} & P & R & F1 & P & R & F1 & P & R & F1 & P & R & F1 & P & R & F1 & P & R & F1 & F1\\
        \midrule
        \textsc{Supervised} & -- & -- & 42.0 & -- & -- & 88.4 & -- & -- & 49.8 & -- & -- & 51.2 & -- & -- & 62.6 & -- & -- & 92.2 & 64.4\\
        \midrule
        \textsc{Extract-Style} &  &  &  &  &  &  &  &  &  &  &  &  &  &  &  &  &  &  & \\
        \hspace{0.5em}zero-shot+\textit{specific} & 0.0 & 0.0 & 0.0 & 60.8 & 12.7 & 21.0 & 16.7 & 15.2 & 15.9 & 17.9 & 2.5 & 4.4 & 9.4 & 4.9 & 6.5 & 24.0 & 7.6 & 11.6 & 9.9\\
        \hspace{0.5em}5-shot & 29.8 & 27.6 & 28.7 & 18.3 & 55.1 & 27.5 & 14.8 & \textbf{40.0} & 21.6 & 26.3 & 22.5 & 24.2 & \textbf{18.6} & \textbf{38.2} & \textbf{25.0} & 29.2 & 29.5 & 29.3 & 26.1\\
        \hspace{1em}\textit{+$k$NN} & 35.3 & 33.3 & 34.3 & 19.0 & 61.8 & 29.0 & \textbf{16.2} & 36.2 & \textbf{22.4} & 29.0 & 29.0 & 29.0 & 15.0 & 34.6 & 20.9 & 27.8 & 41.6 & 33.3 & 28.2\\
        \hspace{1em}\textit{+$k$NN+specific} & 36.4 & 35.0 & 35.7 & 29.6 & 64.3 & 40.5 & 15.4 & 32.4 & 20.9 & 28.0 & 28.7 & 28.4 & 14.4 & 34.6 & 20.3 & \textbf{31.4} & \textbf{51.4} & \textbf{39.0} & \textbf{30.8}\\
        \midrule
        \textsc{Ner-style} &  &  &  &  &  &  &  &  &  &  &  &  &  &  &  &  &  &  & \\
        \hspace{0.5em}zero-shot+\textit{specific} & 22.2 & 1.6 & 3.0 & 12.7 & 5.2 & 7.4 & 2.2 & 1.2 & 1.6 & 2.9 & 0.4 & 0.7 & 2.5 & 2.1 & 2.3 & 0.6 & 0.3 & 0.4 & 2.6\\
        \hspace{0.5em}5-shot & 35.6 & 31.3 & 33.3 & 23.8 & 54.3 & 33.1 & 14.8 & 32.5 & 20.4 & 30.6 & 27.0 & 28.7 & 11.8 & 36.3 & 17.8 & 23.5 & 31.5 & 27.0 & 26.7\\
        \hspace{1em}\textit{+$k$NN} & 36.0 & 37.4 & 36.7 & 21.6 & 63.8 & 32.3 & 10.3 & 30.0 & 15.3 & \textbf{31.0} & \textbf{33.0} & \textbf{32.0} & 10.4 & 30.2 & 15.5 & 26.3 & 41.7 & 32.3 & 27.4\\
        \hspace{1em}\textit{+$k$NN+specific} & \textbf{43.7} & \textbf{44.7} & \textbf{44.2} & \textbf{29.9} & \textbf{64.8} & \textbf{40.9} & 11.2 & 29.2 & 16.1 & 30.8 & 32.8 & 31.8 & 9.1 & 26.6 & 13.7 & 29.8 & 47.3 & 36.6 & 30.6\\
        \midrule
        \textsc{GPT-4*} &  &  &  &  &  &  &  &  &  &  &  &  &  &  &  &  &  &   & \\
        \hspace{0.5em}\textsc{Extract-style} & 40.7 & 35.8 & 38.0 & 52.4 & 66.7 & 58.7 & 19.7 & 35.2 & 25.3 & 31.1 & 30.0 & 30.6 & 20.8 & 41.8 & 27.8 & 31.9 & 55.4 & 40.5  & 36.8\\
        \hspace{0.5em}\textsc{Ner-style} & 50.9 & 45.5 & 48.0 & 69.0 & 66.7 & 67.8 & 19.4 & 33.3 & 24.6 & 24.2 & 20.0 & 21.9 & 20.4 & 34.5 & 25.7 & 29.0 & 56.4 & 38.4  & 37.7\\
        \midrule
        \midrule
        \textsc{RELAX} & P & R & F1 & P & R & F1 & P & R & F1 & P & R & F1 & P & R & F1 & P & R & F1  & F1\\
        \midrule
        \textsc{Extract-Style} &  &  &  &  &  &  &  &  &  &  &  &  &  &  &  &  &  &   & \\
        \hspace{0.5em}5-shot & 72.8 & 67.5 & 70.0 & 23.4 & 70.1 & 35.0 & 31.7 & \textbf{85.7} & \textbf{46.3} & \textbf{76.1} & 65.0 & 70.1 & \textbf{34.0} & 69.8 & 45.7 & \textbf{48.7} & 49.1 & 48.9  & 52.7 \\
        \hspace{1em}\textit{+$k$NN} & \textbf{82.8} & 78.0 & 80.3 & 23.2 & 75.7 & 35.6 & \textbf{33.2} & 74.3 & 45.9 & 74.0 & 74.0 & \textbf{74.0} & 33.1 & 76.8 & \textbf{46.2} & 42.2 & 63.3 & 50.6  & 55.4\\
        \hspace{1em}\textit{+$k$NN+specific} & 80.5 & 77.2 & 78.8 & 34.9 & 75.9 & 47.8 & 33.0 & 69.5 & 44.8 & 72.8 & \textbf{74.4 }& 73.6 & 32.1 & \textbf{77.4} & 45.4 & 46.5 & \textbf{76.0} & \textbf{57.7}  & \textbf{58.0}\\
        \midrule
        \textsc{Ner-style} &  &  &  &  &  &  &  &  &  &  &  &  &  &  &  &  &  &   & \\
        \hspace{0.5em}5-shot & 61.4 & 53.9 & 57.4 & 28.0 & 63.9 & 28.9 & 26.9 & 59.0 & 37.0 & 68.4 & 60.3 & 64.1 & 21.6 & 66.6 & 32.6 & 41.5 & 55.7 & 47.6  & 44.6\\
        \hspace{1em}\textit{+$k$NN} & 80.5 & \textbf{83.7} & \textbf{82.0} & 25.4 & 74.8 & 37.9 & 22.3 & 65.0 & 33.2 & 62.7 & 66.8 & 64.7 & 20.1 & 58.5 & 29.9 & 39.3 & 62.3 & 48.2  & 49.3\\
        \hspace{1em}\textit{+$k$NN+specific} & 79.4 & 81.3 & 80.3 & \textbf{35.2} &\textbf{ 76.3} & \textbf{48.2} & 33.0 & 69.6 & 44.8 & 64.0 & 68.1 & 66.0 & 18.6 & 53.9 & 27.6 &  42.0 & 66.6 & 51.4  & 53.1\\
        \bottomrule
    \end{tabular}}
    \caption{\textbf{Results of Experiments.} The metrics are Precision (\textbf{P}), Recall (\textbf{R}) and span-F1 (\textbf{F1}). \textit{*For GPT-4, the results are based on a subset of  $\leq350$ samples for each dataset.} 
    }
    \label{tab:results-full}
\end{table*}

\begin{figure}[ht!]
  \centering
  \includegraphics[width=0.4\textwidth]{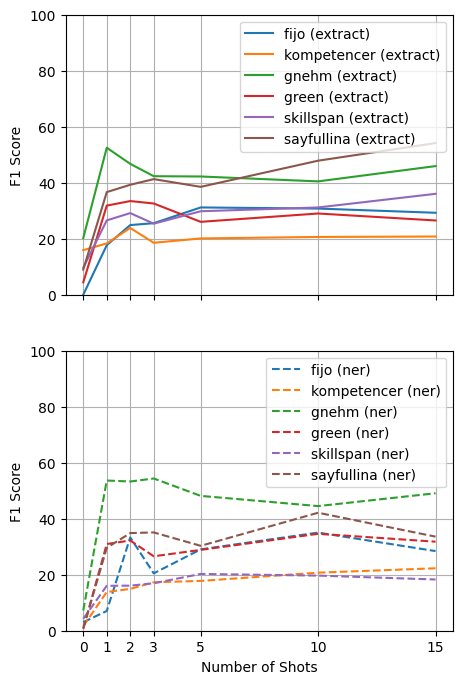}
  \caption{F1 performances of \textsc{Extract-style} and \textsc{Ner-style} on the full dataset of \textsc{Fijo}, \textsc{Kompetencer}, \textsc{Green}, and subset of 350 samples from \textsc{Gnehm}, \textsc{Skillspan}, and \textsc{Sayfullina}. We recorded the experiments using $k=\{0, 1, 2, 3, 5, 10, 15\}$ shots. Demonstrations are retrieved randomly.}
  \label{fig:app-fewshots}
\end{figure}

\begin{figure}[ht!]
  \centering
  \includegraphics[width=0.4\textwidth]{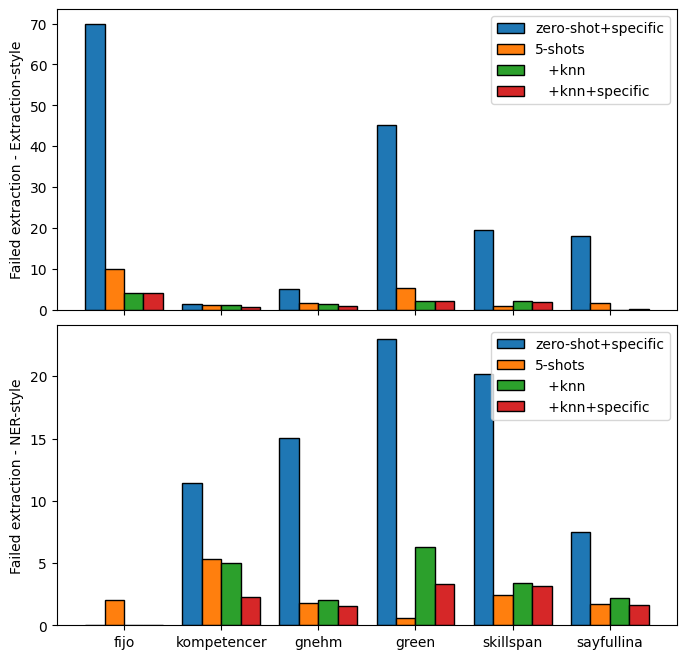}
  \caption{Percentage of samples in which LLM failed to extract entities after 3 re-tries. The zero-shot setting gave the most number of failures.}
  \label{fig:failures}
\end{figure}

\paragraph{Impact of the number of demonstrations}

We conducted experiments to examine the impacts of the number of demonstrations $k$ on the extraction results. Experiments are done on the full test set of \textsc{Fijo}, \textsc{Kompetencer}, \textsc{Green}, and a subset of 350 test samples for the other datasets to avoid an overhead in experimental cost. The result are recorded in Figure \ref{fig:failures}. Surprisingly, unlike the regarded knowledge that more demonstrations lead to better predictive capabilities, we only observed incremental performances for the first few demonstrations. After $k=3$, we found that adding demonstrations did not guarantee better extraction results. Even for smaller $k$, certain datasets such as \textsc{Gnehm} also exhibited irregular performance patterns. We also found that increases are more stable for \textsc{Ner-style} approach, partially due to the highly structured and specific output format it requires. \\
Note that in our experiments, $k=i$ means that there are $i$ positive and $i$ negative examples in the demonstrations.

\paragraph{Failure cases}
Further breakdown of failure cases for each dataset can be found in Figure \ref{fig:failures}. In general, we found that the failure rates highly vary between datasets.

\paragraph{Impact of negative demonstrations}
Table \ref{tab:app-positive_impact} compares providing 10 positive examples as demonstrations with mixing 5 positive and 5 negative demonstrations. Experiments are done in a subset of $\leq$ 350 test samples with 5 shots.

\begin{table}[!ht]
\centering
\resizebox{\linewidth}{!}{%
\begin{tabular}{llcc}
\toprule
\multicolumn{2}{c}{\textbf{\textsc{Dataset}}} &\textbf{\textsc{Extract-style}} & \textbf{\textsc{Ner-style}} \\
\midrule
\multirow{ 2}{*}{\textsc{Fijo}} & \textbf{Positive} & 28.2 / 6 / 2 & 28.0 / 3 / 8 \\
& \textbf{Mix} & \textbf{31.2} / 6 / 4 & \textbf{29.1} / 5 / 7 \\
\multirow{ 2}{*}{\textsc{Kompetencer}} & \textbf{Positive} & \textbf{22.9} / 129 / 2 & \textbf{18.2} / 103 / 4 \\
& \textbf{Mix} & 20.1 / 131 / 1 & 17.8 / 133 / 12 \\
\multirow{ 2}{*}{\textsc{Gnehm}} & \textbf{Positive} & \textbf{43.4} / 219 / 11 & 44.8 / 174 / 11 \\
& \textbf{Mix} & 42.2 / 210 / 6 & \textbf{54.4} / 215 / 17 \\
\multirow{ 2}{*}{\textsc{Green}} & \textbf{Positive} & \textbf{34.3} / 69 / 21 & \textbf{35.0} / 17 / 10 \\
& \textbf{Mix} & 26.0 / 62 / 17 & 28.9 / 47 / 17 \\
\multirow{ 2}{*}{\textsc{Skillspan}} & \textbf{Positive} & \textbf{31.0} / 212 / 3 & 17.9 / 125 / 5 \\
& \textbf{Mix} & 29.8 / 245 / 7 & \textbf{20.3} / 183 / 10 \\
 \multirow{ 2}{*}{\textsc{Sayfullina}} & \textbf{Positive} & \textbf{52.2} / 1 / 72 & \textbf{38.8} / 0 / 36 \\
 & \textbf{Mix} & 38.5 / 1 / 87 & 30.3 / 0 / 51 \\
 \midrule
 \multirow{ 2}{*}{\textsc{Avg}} & \textbf{Positive} & \textbf{35.3} & 30.5\\
& \textbf{Mix} & 31.3 & 30.3 \\
\bottomrule
\end{tabular}%
}
\caption{Experiments using positive-only demonstrations vs. a mixture of positive-negative. We report \textbf{F1 / TN / FN}, with \textbf{TN / FN} being the number of true and false negative predictions LLM made (cases where LLM predicted there were no entities).}
\label{tab:app-positive_impact}
\end{table}

For \textsc{Extract-style} prompting, providing positive-only demonstrations improves the F1 score for all datasets. Besides, positive-only context does not affect the frequency to which LLM predicts an example as negative (have \textbf{None} entities). On the other hand, the results vary for \textsc{Ner-style}. Notably, for this strategy, using positive-only demonstrations significantly abstains LLM from making negative predictions, with an average decrease of \textbf{None} samples of 44.9\% and 29.1\% for 3-shot and 5-shot, respectively.  By allowing \textbf{None} predictions, a mixture of positives and negatives increases the \textbf{TN} performances for \textsc{Ner-style}, while only slightly increasing the number of \textbf{FN}. This phenomenon, perhaps, is due to the fact the instruction from \textsc{Extract-style} is obvious and straightforward, from which the LLM can infer the ability to label \textbf{None} directly. Therefore, providing LLM with more positive examples solidifies the semantic understanding of the skills extraction task the LLM is solving. Meanwhile, \textsc{Ner-style} is very specific in its output and therefore the LLM tends to over-generalize the positive-only NER-format sentences it learns from the in-context demonstrations. Besides, mix demonstrations also results in comparable number of \textbf{None} predictions for both \textsc{Extract-style} and \textsc{Ner-style}, thus we utilized this approach for our experiments.

\begin{table*}[!ht]
\centering
\resizebox{\linewidth}{!}{
\begin{tabular}{lll}
\toprule
\textbf{\textsc{Dataset}} & &\textbf{\textsc{Prompt}}  \\
\midrule
\multirow{ 3}{*}{\textbf{\textsc{Fijo}}} 
& \textsc{System} & \parbox[t]{15cm}{You are an expert human resource manager in the insurance industry in France. You need to analyse skills required in job offers.}  \\
& \textsc{Extract-style} & \parbox[t]{15cm}{You are given a sentence from an insurance job description in French. Extract all the skills and competencies that are required from the candidate as list, with one skill per line. If no skill is found in the sentence, return "None".} \\
& \textsc{Ner-style} & \parbox[t]{15cm}{You are given a sentence from an insurance job description in French. Highlight all the skills and competencies that are required from the candidate, by surrounding them with tags '@@' and '\#\#'. If there are no such element in the sentence, replicate the sentence identically.} \\
\midrule
\multirow{ 3}{*}{\textbf{\textsc{Kompetencer}}} 
& \textsc{System} & \parbox[t]{15cm}{You are an expert human resource manager. You need to analyse skills required in job offers.}  \\
& \textsc{Extract-style} & \parbox[t]{15cm}{You are given a sentence from a job description in Danish. Extract all the skills, knowledges, and competencies that are required from the candidate as list, with one skill per line. If no skill is found in the sentence, return "None".} \\
& \textsc{Ner-style} & \parbox[t]{15cm}{You are given a sentence from a job description in Danish. Highlight all the skills, knowledges, and competencies that are required from the candidate, by surrounding them with tags '@@' and '\#\#'. If there are no such element in the sentence, replicate the sentence identically.} \\
\midrule
\multirow{ 3}{*}{\textbf{\textsc{Gnehm}}} 
& \textsc{System} & \parbox[t]{15cm}{You are an expert human resource manager in information and communication technology (ICT) from Germany. You need to analyse skills required in German job offers.}  \\
& \textsc{Extract-style} & \parbox[t]{15cm}{You are given a sentence from a job advertisement in German. Extract all the IT/Technology skills and competencies that are required from the candidate as list, with one skill per line. If no skill is found in the sentence, return "None".} \\
& \textsc{Ner-style} & \parbox[t]{15cm}{You are given an extract from a job advertisement in German. Highlight all the IT/Technology skills and competencies that are required from the candidate, by surrounding them with tags '@@' and '\#\#'. If there are no such element in the sentence, replicate the sentence identically.} \\
\midrule
\multirow{ 3}{*}{\textbf{\textsc{Green}}} 
& \textsc{System} & \parbox[t]{15cm}{You are an expert human resource manager. You need to analyse skills required in job offers.}  \\
& \textsc{Extract-style} & \parbox[t]{15cm}{You are given a sentence from a job descriptionin various fields like IT, finance, healthcare, and sales. Extract all the skills and competencies that are required from the candidate as list, with one skill per line. If no skill is found in the sentence, return "None"} \\
& \textsc{Ner-style} & \parbox[t]{15cm}{You are given a sentence from a job description in various fields like IT, finance, healthcare, and sales. Highlight all the skills and competencies that are required from the candidate, by surrounding them with tags '@@' and '\#\#'.  If there are no such element in the sentence, replicate the sentence identically.} \\
\midrule
\multirow{ 3}{*}{\textbf{\textsc{Skillspan}}} 
& \textsc{System} & \parbox[t]{15cm}{You are an expert human resource manager. You need to analyse skills required in job offers.}  \\
& \textsc{Extract-style} & \parbox[t]{15cm}{You are given a sentence from a job posting. Extract all the skills, knowledges, and competencies that are required from the candidate as list, with one skill per line. If no skill is found in the sentence, return "None".} \\
& \textsc{Ner-style} & \parbox[t]{15cm}{You are given a sentence from a job posting. Highlight all the skills, knowledges, and competencies that are required from the candidate, by surrounding them with tags '@@' and '\#\#'. If there are no such element in the sentence, replicate the sentence identically.} \\
\midrule
\multirow{ 3}{*}{\textbf{\textsc{Sayfullina}}} 
& \textsc{System} & \parbox[t]{15cm}{You are an expert human resource manager. You need to detect and analyse soft skills required in job offers.}  \\
& \textsc{Extract-style} & \parbox[t]{15cm}{You are given a sentence from a job advertisement. Extract all the soft skills and competencies that are required from the candidate as list, with one skill per line. If no skill is found in the sentence, return "None"} \\
& \textsc{Ner-style} & \parbox[t]{15cm}{You are given a sentence from a job advertisement. Highlight all the soft skills and competencies that are required from the candidate, by surrounding them with tags '@@' and '\#\#'. If there are no such element in the sentence, replicate the sentence identically.} \\
\bottomrule
\end{tabular}}
\caption{Full \textit{data-specific} prompts for each dataset.}
\label{tab:prompt}
\end{table*}
\section{Specific Prompts used for each dataset}
\label{app:prompting}

The instructions we use for prompting LLMs for each dataset are in Table \ref{tab:prompt}.

\section{Mismatches types and examples}
\label{ap:mismatches}
We provide some examples of common mismatches from skipping/extra spaces and punctuation insertion, as well as examples where the mismatches are based on the pre-trained knowledge of LLMs that can not be modified even with explicit feedback prompt guidance.

\paragraph{Successful sample with minor edits.} Minor sentence construction (with regards to punctuations, spaces, and simple grammar errors) that was addressed and fixed by LLMs. Here are a few examples.

\textit{\textbf{Spaces and punctuations:}}

\textit{Original:} Test Consultant / Automation Test Analyst will ideally be confident with Selenium and good experience of web based testing , HTML and JavaScript .

\textit{Generation:} @@Test Consultant/Automation Test Analyst\#\# will ideally be confident with @@Selenium\#\# and good experience of @@web based testing\#\#, @@HTML\#\# and @@JavaScript\#\#.

\textit{\textbf{Fixable grammar and vocabulary mistakes:}}

\textit{Original:} Sevenoaks , Kent Support Engineer with strong knowledge of Windows and Linux architecture is required by an expanding company based in Sevenoaks to support clients throughout London and Kent .

\textit{Generation:} @@Support Engineer@@ with @@strong knowledge\#\# of @@Windows\#\# and @@Linux architecture\#\# is required by an expanding company based in Sevenoaks to support clients throughout London and Kent .

\textit{Fixed:} Sevenoaks , Kent Support Engineer with strong knowledge of @@Windows\#\# and @@Linux architecture\#\# is required by an expanding company based in Sevenoaks to support clients throughout London and Kent .

\paragraph{Failed sample due to common grammar knowledge.} LLMs failed to output the exact original sentence because they tend to "over fix" commonly known grammar mistakes. For example, LLMs cannot adhere to the original sentence in the following input, which lacks "To" in "To be considered", even after 3 rounds of feedback loops.

\textit{Original:} be considered for the General Manager post you will have at least 2years management experience within the health and fitness industry ideally as a General Manager . 

\textit{Generation:} To be considered for the @@General Manager\#\# post you will have at least 2years management experience within the @@health and fitness industry\#\# ideally as a @@General Manager\#\# .

\section{Qualitative Analysis}
\label{ap:qual_analysis}
We provide examples from datasets \textsc{Green}, \textsc{Skillspan}, \textsc{Sayfullina}, and \textsc{Fijo} that we believe are more representative of the 60 examples we examined for qualitative analysis in Table \ref{tab:qual_analysis}. We note that this study is highly subjective and does not conform to any specific definition of error categories.

\begin{table*}[!ht]
    \centering
    \resizebox{\linewidth}{!}{
    \begin{tabular}{llllll}
    \toprule
    \toprule
        Error Type & Dataset Name        & Sentence  & Labels   & \textsc{Extract-style}  & \textsc{Ner-style}  \\
        \midrule
        \midrule
        \parbox[t]{3cm}{Skill definition mis-alignment} & \textsc{Sayfullina}   & \parbox[t]{7cm}{creative and confident when it come to meeting deadlines this could be the} & 
        \parbox[t]{4cm}{meeting deadlines} & 
        \parbox[t]{4cm}{creative, confident} & 
        \parbox[t]{7cm}{creative and confident when it come to meeting deadlines this could be the} \\
        \midrule

        \parbox[t]{3cm}{Skill definition mis-alignment} & \textsc{Skillspan}   & \parbox[t]{7cm}{Design and build front-end / BI layer of our data and analytics solutions - Power BI reports DAX queries Azure Analysis Services models .} & 
        \parbox[t]{4cm}{Design and build front-end / BI layer} & 
        \parbox[t]{4cm}{Design and build front-end / BI layer of our data and analytics solutions, Power BI reports, DAX queries, Azure Analysis Services models} & 
        \parbox[t]{7cm}{Design and build front-end / BI layer of our data and analytics solutions - Power BI reports DAX queries Azure Analysis Services models .} \\
        \midrule
        \parbox[t]{3cm}{Skill definition mis-alignment} & \textsc{Sayfullina}   & \parbox[t]{7cm}{focus , well organize with excellent communication and organizational skills} & 
        \parbox[t]{4cm}{organizational skills} & 
        \parbox[t]{4cm}{focus, organizational skills} & 
        \parbox[t]{7cm}{focus , well organize with excellent @@communication\#\# and @@organizational\#\# skills} \\
        \midrule
        \parbox[t]{3cm}{Conjoined skill} & \textsc{Skillspan}  &  
        \parbox[t]{7cm}{Optimize and manage existing reward and benefits programmes and processes to support the business priorities and ensure local compliance.} & 
        \parbox[t]{4cm}{Optimize and manage existing reward and benefits programmes and processes, support the business priorities, ensure local compliance} &  
        \parbox[t]{4cm}{optimize reward and benefits programmes, manage reward and benefits processes, support business priorities, ensure local compliance} & 
        \parbox[t]{7cm}{@@Optimize and manage existing reward and benefits programmes and processes\#\# to @@support the business priorities\#\# and @@ensure local compliance\#\# .} \\
        \midrule
        \parbox[t]{3cm}{Conjoined skill} & \textsc{Green}   & \parbox[t]{7cm}{leading manufacturing business who supply luxury products to the Retail and Leisure industries are looking to recruit a talented Supply Chain Manager to manage operations , production , warehousing , delivery , reporting and to develop best practice Supply Chain procedures within the organisation .} & 
        \parbox[t]{4cm}{delivery, develop best practice Supply Chain procedures, warehousing, manage operations, reporting, production} & 
        \parbox[t]{4cm}{Supply Chain Manager, operations, production, warehousing, delivery, reporting, develop best practice Supply Chain procedures} & 
        \parbox[t]{7cm}{leading manufacturing business who supply luxury products to the Retail and Leisure industries are looking to recruit a talented Supply Chain Manager to @@manage operations\#\# , @@production\#\# , @@warehousing\#\# , @@delivery\#\# , @@reporting\#\# and to @@develop best practice Supply Chain procedures\#\# within the organisation .} \\

        \midrule
        \parbox[t]{3cm}{Extended Span} & \textsc{Green}   & \parbox[t]{7cm}{Frank International is the leading Microsoft Dynamics recruitment firm in the UK , advertising more Dynamics CRM jobs than any other agency .} & 
        \parbox[t]{4cm}{Microsoft Dynamics, Dynamics CRM} & 
        \parbox[t]{4cm}{Microsoft Dynamics recruitment, Dynamics CRM jobs} & 
        \parbox[t]{7cm}{Frank International is the leading @@Microsoft Dynamics recruitment firm\#\# in the UK , advertising more @@Dynamics CRM jobs\#\# than any other agency .} \\
        \midrule
        \parbox[t]{3cm}{Extended Span} & \textsc{Green}   & \parbox[t]{7cm}{to degree level ideally in business strong relationship builder organise and analytical polish} & 
        \parbox[t]{4cm}{relationship builder} & 
        \parbox[t]{4cm}{strong relationship builder, organise, analytical} & 
        \parbox[t]{7cm}{to degree level ideally in business @@strong relationship builder\#\# organise and analytical polish} \\
        \midrule
        \parbox[t]{3cm}{Wrong annotations} & \textsc{Green}   & \parbox[t]{7cm}{will still be part of an office team with all the community and collective achievement that goes with that . This} & 
        \parbox[t]{4cm}{be part of an office team} & 
        \parbox[t]{4cm}{None} & 
        \parbox[t]{7cm}{will still be part of an office team with all the @@community\#\# and @@collective achievement\#\# that goes with that . This} \\
        \midrule
        \parbox[t]{3cm}{Wrong annotations} & \textsc{Sayfullina}   & \parbox[t]{7cm}{any offer be subject to rigorous reference , criminal and credit check} & 
        \parbox[t]{4cm}{rigorous} & 
        \parbox[t]{4cm}{rigorous reference, credit check} & 
        \parbox[t]{7cm}{any offer be subject to rigorous @@reference\#\# , @@criminal\#\# and @@credit check\#\#} \\
        \midrule
        \parbox[t]{3cm}{Wrong extraction} & \textsc{Skillspan}   & \parbox[t]{7cm}{You will be a part of a small team of 4 people in the division who closely cooperate with our <LOCATION> <LOCATION> office in the development process .} & 
        \parbox[t]{4cm}{} & 
        \parbox[t]{4cm}{cooperate with our <LOCATION> <LOCATION> office} & 
        \parbox[t]{7cm}{You will be a part of a small team of 4 people in the division who closely cooperate with our @@<LOCATION> <LOCATION> office\#\# in the development process .} \\
        \midrule
        \parbox[t]{3cm}{Wrong extraction} & \textsc{Sayfullina}   & \parbox[t]{7cm}{be willing to travel solo internationally infrequently have} & 
        \parbox[t]{4cm}{willing to travel} & 
        \parbox[t]{4cm}{None} & 
        \parbox[t]{7cm}{be willing to travel solo internationally infrequently @@have\#\#} \\
        \bottomrule
        \bottomrule
    \end{tabular}}
    \caption{Qualitative examples failed extractions for each category. LLM sometimes cannot conform to the strict NER-output requirements, even though its extraction makes sense.}
    \label{tab:qual_analysis}
\end{table*}

\begin{table*}[!ht]
    \centering
    \begin{tabular}{llrr}
    \toprule
    Prompt & Dataset & Span Length Difference & Number of Skills Difference \\
    \midrule
    \multirow{6}{*}{\textsc{Ner}} & \textsc{green} & 2.15 & -0.03 \\
     & \textsc{skillspan} & 2.32 & -0.49 \\
     & \textsc{fijo} & -2.25 & -0.06 \\
     & \textsc{sayfullina} & 3.54 & -0.56 \\
     & \textsc{kompetencer} & 3.12 & -0.49 \\
     & \textsc{gnehm} & 3.72 & -0.36 \\
    \midrule
    \textsc{Ner} & avg & 2.10 & -0.33 \\
    \midrule
    \multirow{6}{*}{\textsc{Extract}} & green & 1.28 & -0.04 \\
     & \textsc{skillspan} & 1.31 & -0.43 \\
     & \textsc{fijo} & -2.81 & 0.10 \\
     & \textsc{sayfullina} & 3.32 & -0.64 \\
     & \textsc{kompetencer} & 2.03 & -0.44 \\
     & \textsc{gnehm} & 3.64 & -0.41 \\
    \midrule
    \textsc{Extract} & avg & 1.46 & -0.31 \\
    \bottomrule
\end{tabular}
    \caption{Difference between skills extracted by GPT-3.5 and gold annotations, for each dataset and prompting strategy. We use $k$NN-5-shots with dataset-specific prompts.}
    \label{tab:app-differences}
\end{table*}

\end{document}